\documentclass[11pt]{article}
\newcommand{\tab}{\hspace*{2em}}

\usepackage[preprint]{acl}

\usepackage{times}
\usepackage{latexsym}

\usepackage[T1]{fontenc}
\usepackage[utf8]{inputenc}

\usepackage{microtype}

\usepackage{inconsolata}

\usepackage{graphicx}

\title{Rule-Based Approaches to Atomic Sentence Extraction}

\author{Lineesha Kamana \\
  The Pennsylvania State University \\
  \texttt{lpk5305@psu.edu} \\\And
  Akshita Ananda Subramanian \\
  The Pennsylvania State University \\
  \texttt{apa6321@psu.edu} \\
  \AND
  Mehuli Ghosh \\
  The Pennsylvania State University \\
  \texttt{mvg6189@psu.edu} \\\And
  Suman Saha \\
  The Pennsylvania State University \\
  \texttt{szs339@psu.edu} \\}

\begin{document}
\maketitle
\begin{abstract}
\tab Natural language often combines multiple ideas into complex sentences. Atomic sentence extraction, the task of decomposing complex sentences into simpler sentences that each express a single idea, improves performance in information retrieval, question answering, and automated reasoning systems. Previous work has formalized the “split-and-rephrase” task and established evaluation metrics, and machine learning approaches using large language models have improved extraction accuracy. However, these methods lack interpretability and provide limited insight into which linguistic structures cause extraction failures. Although some studies have explored dependency-based extraction of subject-verb-object triples and clauses, no principled analysis has examined which specific clause structures and dependencies lead to extraction difficulties.

\tab This study addresses this gap by analyzing how complex sentence structures—including relative clauses, adverbial clauses, coordination patterns, and passive constructions—affect rule-based atomic sentence extraction performance. Using the WikiSplit dataset, we implemented dependency-based extraction rules in spaCy, generated 100 gold-standard atomic sentence sets, and evaluated performance using ROUGE and BERTScore. The system achieved \textbf{ROUGE-1 F1 = 0.6714}, \textbf{ROUGE-2 F1 = 0.478}, \textbf{ROUGE-L F1 = 0.650}, and \textbf{BERTScore F1 = 0.5898}, indicating moderate to high lexical, structural, and semantic alignment. Challenging structures included relative clauses, appositions, coordinated predicates, adverbial clauses, and passive constructions. Overall, rule-based extraction is reasonably accurate but sensitive to syntactic complexity.

\end{abstract}

\section{Introduction}

\tab Many natural language sentences contain multiple propositions expressed through embedded clauses and other complex linguistic structures. For tasks such as question answering and automated reasoning, these sentences can contain too many ideas packed together, which makes it harder for computer systems to understand the meaning clearly. This can make them perform worse on such tasks. 

Atomic sentence extraction can help bypass this challenge. It unpacks complex sentences into simple substances that each convey a single idea. These smaller “atomic” units are easier for systems to interpret as they make relationships between subjects, verbs, and prediction more clear. 

Previous work has explored this sector through the split-and-rephrase problem and relies heavily on large neural models trained on datasets. While these models perform well, they do not explain why they choose certain splits, and it is hard to understand which linguistic structures cause them to fail. Rule-based and dependency-based approaches, on the other hand, provide far more transparency as they rely on the syntax itself.

To explore which specific linguistic structures cause the most problems for rule-based atomic sentence extraction systems, two research questions will be answered:\\
\begin{center}
\textbf{What types of sentence structures (such as relative clauses, conditionals, or negations) are the hardest to split correctly?}\\

\end{center}
\begin{center}
\textbf{How accurate are the sentence extractions?}\\
\end{center}

To answer these questions, we built a rule-based extraction system using spaCy’s dependency parser and evaluated it on a subset of the WikiSplit dataset. The goal of this is to pinpoint which features lead to errors and determine how future systems can better handle even more complex sentences.

\section{Background and Literature Review}

\tab Research on splitting up complex sentences into simpler, atomic units, emerged from the split-and-rephrase task. Narayan et al. (2017) formalized this approach. Their work introduced evaluation metrics such as BLEU and SARI for measuring accuracy and usefulness of the split sentences. This has established a strong foundation for future studies then adopted. 

Building on this framework, Botha et al. (2018) introduced WikiSplit, which is a large dataset that was created from naturally occurring sentence splits found in Wikipedia edit histories. This work has helped demonstrate that a large-scale dataset can significantly improve AI models for the split-and-rephrase approach. The WikiSplit dataset also offered a more diverse set of sentence structures compared to datasets used in previous work. This makes it a very valuable benchmark for evaluating real-world splitting. 

In comparison to a neural network approach, dependency-based extraction methods focus on the interpretability of the sentence structures’ syntax. Nikalus et al. (2018) provided a survey of Open Information Extraction (OpenIE) systems and highlighted how dependency parses can be used to extract subject-verb-object tropes and clause-like units. The discussion of static relations, such as nsubj, dobj, advcl, and elcl directly influences how we formulated rules for identifying atomic sentences. Their survey also emphasized the strengths and limitations of rule-based systems, which reinforced the need for detailed analysis of which linguistic structures cause problems for dependency-based extraction. 

Syntactic simplification research has shown that reducing grammatical complexity by operations such as disembedding relative clauses, separating subordinate clauses, and splitting coordinated verb phrases can substantially improve text comprehensibility for readers and downstream NLP systems (Siddharthan, 2006).

Liu, Cohen, and Lapata (2018) introduced the methods for parsing Discourse Representation Structures (DRS). This has emphasized interpretable and rule-based derivations. Their work demonstrated the value behind structured representations for semantic sentences. This has supported our decision to build a rule-based extraction model rather than a neural model, ensuring we can trace back how specific syntax influences the extraction success. 

These studies provide the conceptual and methodological abscess for our work. Prior research had helped to identify datasets and evaluation frameworks as well as demonstrating the usefulness of dependency information and highlighting the importance of interpretability. However, none of these studies examine which clause types cause challenges with rule-based atomic sentence extraction. 

\section{Methodology}

\subsection{Dataset Selection}

\tab We sampled 100 complex sentences from the WikiSplit dataset, which is a large-scale corpus built for the purpose of the split-and-rephrase task. WikiSplit, however, only provides two-way splits, which are insufficient for evaluating the success of multi-clause atomic decomposition. Because our goal was to break up a sentence into all propositions that express a single idea, we treated the complex sentences as a base for developing more granular atomic sentences.

\subsection{Gold-Standard Annotation Procedure}

\tab Each of the 100 sentences that were selected were manually split into 2-5 atomic sentences, depending on the number of separate ideas in the overall complex sentence. During annotation, guidelines were strictly followed to ensure that splitting was consistent across all sentences.

\begin{itemize}
    \item Each atomic sentence must express only one state or event.
    \item Relative clauses such as (“Diana, who served as mayor…”) were split into separate atoms.
    \item Coordinated predicates such as (“Alic sang and danced”) were split into multiple sentences, such as “Alice sang” and “Alice danced.”
    \item Adverbial clauses such as “before” and “although” were split into their own separate sentences.
\end{itemize}

\subsection{Data Cleaning and Tokenization}

\tab To prepare the dataset for splitting, and eventually evaluation, duplicate sentences were first removed using Pandas. Then, all sentences were made lowercase for matching, so that lack of capitalization would not skew the original sentence splitting rules, and later on, the ROUGE or BERT Scores. Then, spaCy’s “en\_core\_web\_sm” model was used because it is efficient and lightweight. It was used to tokenize, tag for part of speech, and parse for dependency.

\subsection{Rules-Based Splitting}

\tab A rules-based extraction system attempted to split each sentence by detecting coordinating conjunctions using the words themselves (“and” and “or”), or the part of speech (“conj”). It also identified clauses that were tagged as being subordinate, such as adverbial and relative clauses (“advcl” and “relcl”). It also attempted to, based on dependency parsing, assign the correct subject to a phrase.

\begin{quote}
“Anna ate an apple and a banana” → “Anna ate an apple,” “Anna ate a banana”  
\end{quote}
The system’s coordinating conjunction detection, in this case, eliminates “and” from either sentence and creates a separate “ate” for “a banana”, and subject reconstruction assigns “Anna” to “ate a banana.”

\subsection{Evaluation Against Gold Standard Data}

\tab The atomic sentences generated by the model were paired with their equivalents in the manually created gold standard set. To compute the accuracy of the generated atomic sentences, the following metrics were calculated:

\begin{itemize}
    \item ROUGE-1, ROUGE-2, ROUGE-L, which quantify the lexical and structural overlaps between model-produced atomic sentences and gold standard data.
    \item BERTScore (using RoBERTa-large), which shows semantic similarities between model-produced atomic sentences and gold standard data.
\end{itemize}

\subsection{Error Categories}

\tab Additionally, after observation of errors across the set of model-produced atomic sentences, we identified multiple areas where the rules-based system failed, along with an example in each category from the data.\\

\subsubsection*{Missing Subject}
\begin{quote}
\textit{Model output}: “Renamed him Wei Zhongxian.”\\
\textit{Gold}: “The couple renamed him Wei Zhongxian.”\\
\textbf{→} The model starts with a verb phrase and loses the original noun phrase subject.
\end{quote}
\subsubsection*{Missing Object}
\begin{quote}
\textit{Model output}: “Nadal lost in straight sets.”\\
\textit{Gold}: “Nadal lost the match in straight sets.”\\
\textbf{→} The model preserves the event but drops the object “the match,” creating an underspecified predicate.
\end{quote}
\subsubsection*{Coordination Error}
\begin{quote}
\textit{Model output}: “She taught herself how to draw.”\\
\textit{Gold}: “She taught herself to draw and began selling cartoons while still in high school.”\\
→ The coordinated predicate “began selling cartoons” is entirely lost.
\end{quote}
\subsubsection*{Relative Clause Error}
\begin{quote}
\textit{Model output}: “Marks \& Co. was located in the five-story building.”\\
\textit{Gold}: “The five-story building where Marks \& Co. was located still exists.”\\
→ The relative clause “where Marks \& Co. was located” is treated as the main clause, which reverses the meaning.
\end{quote}
\subsubsection*{Adverbial Clause Error}
\begin{quote}
\textit{Model output}: “He played on Broadway.” \\
\textit{Gold}: “During his twenty-year stay in the USA, he played on Broadway.”\\
→ The adverbial modifier “During his twenty-year stay in the USA” is lost.
\end{quote}
\subsubsection*{Appositive Error}
\begin{quote}
\textit{Model output}: “Maria Johanna Görtz was a Swedish artist.”\\
\textit{Gold}: ““Maria Johanna Görtz, also known as Jeanette Görtz, was a Swedish artist.”\\
→ The appositive “also known as Jeanette Görtz” is dropped.
\end{quote}
\subsubsection*{Truncated}
\begin{quote}
\textit{Model output}: “The album was released”\\
\textit{Gold}: “The album was released in limited numbers and only in a 12'' vinyl format.”\\
→ The model only extracts the main verb phrase and truncates the rest of the idea, even when it could possibly be used in another atomic sentence.
\end{quote}
\subsubsection*{Other}
\begin{quote}
\textit{Model output}: “Some icebergs calved and passed out of the fjord.”\\
\textit{Gold}: “Around 10\% of Greenland’s icebergs calve and flow out of the fjord annually.”\\
→ There are multiple issues, including that it is missing a numerical detail, missing the modifier “annually,” and has the wrong tense.
\end{quote}

\section{Results}

\subsection{Overall Performance}

\tab The rule-based atomic sentence extraction system was evaluated against the previously discussed manually annotated gold-standard set of 100 WikiSplit source sentences, which produced 252 aligned atomic sentence pairs. Performance was measured using ROUGE and BERTScore, all of which together work to evaluate the lexical, structural, and semantic similarity of the sentence pairs.\\

\begin{table}[!h]
\small
\centering
\begin{tabular}{lccc}
\hline
\textbf{Metric} & \textbf{Precision} & \textbf{Recall} & \textbf{F1} \\
\hline
ROUGE-1 & 0.7645 & 0.6328 & 0.6714 \\
ROUGE-2 & 0.5595 & 0.4473 & 0.4776 \\
ROUGE-L & 0.7427 & 0.6098 & 0.6495 \\
\hline
\end{tabular}
\caption{ROUGE scores comparing model-generated atomic sentences to the gold standard.}
\label{tab:rouge}
\end{table}
The ROUGE-1 F1 score indicates that the token-level similarity between the gold standard and model-produced score is moderate-to-high, and the ROUGE-L F1 score suggests that the rules-based extraction system preserves the gold standard sentence structure. However, the ROUGE-2 F1 score shows that the rules-based system does poorly in preserving local phrasal similarity.

\begin{table}[!h]
\small
\centering
\begin{tabular}{lccc}
\hline
\textbf{Metric} & \textbf{Precision} & \textbf{Recall} & \textbf{F1} \\
\hline
BERTScore & 0.5898 & 0.5414 & 0.5648 \\
\hline
\end{tabular}
\caption{BERTScore semantic similarity between model and gold atomic sentences.}
\label{tab:bertscore}
\end{table}
The BERTScore is moderate, which indicates that the rules-based extraction moderately preserves semantic similarity, but might fail when working with deeper semantic content. This is typically due to missing arguments or incomplete clause reconstruction.

\subsection{Length and Verb Characteristics}

\tab To help quantify the structural differences between the atomic sentences produced by the rules-based decomposition system and the gold standard, the average token length and number of verbs were considered to see if important structural portions were omitted.

\begin{table}[!h]
\small
\centering
\begin{tabular}{p{0.3\columnwidth}cc}
\hline
\textbf{Measure} & \textbf{Model-Produced} & \textbf{Gold Standard} \\
\hline
Average Atomic \\Sentence Length & 11.3769 & 12.6666 \\
Average Verbs per \\Atomic Sentence & 1.2857 & 1.4285 \\
\hline
\end{tabular}
\caption{Comparison of length and verb count for model-produced and gold-standard atomic sentences.}
\label{tab:lengthverbs}
\end{table}

\subsection{Error Categories}

\tab After performing manual categorization of error types using dependency features, 8 major categories were identified, including one “Other” category which includes multiple types of error in the same atomic sentence.

\begin{table}[!h]
\small
\centering
\begin{tabular}{lc}
\hline
\textbf{Error Type} & \textbf{Proportion of Errors} \\
\hline
Missing Object & 0.444444 \\
Multiple / Other & 0.392857 \\
Coordination Error & 0.051587 \\
Missing Subject & 0.043651 \\
Appositive Error & 0.027778 \\
Adverbial Clause Error & 0.019841 \\
Relative Clause Error & 0.015873 \\
Truncated & 0.003968 \\
\hline
\end{tabular}
\caption{Distribution of error types in model-generated atomic sentences.}
\label{tab:errors}
\end{table}
The most prevalent error category is Missing Object, in which the system correctly identifies a predicate but fails to include the corresponding object. An example of an incomplete sentence produced by the rules-based system that meets this error category is “Lady Yuhwa got,” which lacks an object. The second-most common error category consists of miscellaneous or combinations of incorrect outputs that occur when the rules-based extractor mishandles more complex dependencies. The proportion of this category indicates the pervasive nature of these errors such that many atomic sentences have multiple errors.

Coordination structures can also result in errors when the system either fails to split coordinated predicates, or incorrectly treats coordinated noun phrases as separate events. Missing Subject errors can occur when subject propagation fails across combined clauses; this is more often seen in passive constructions. Additionally, relative, adverbial, and appositive clause errors occur far less frequently, but also reflect the difficulty of identifying clause boundaries. In complex sentences with nested clauses such as in the WikiSplit dataset, syntactic cues for clause boundaries are often ambiguous.

\subsection{Summary of Results}

\tab Overall, the results of this work show that lexical overlap is strong, but grammatical completeness remains inconsistent. As a result, semantic similarity between the atomic sentences generated by the rules-based system and the gold standard sentences is only moderate, which indicates that many predicted atomic sentences preserve the correct meaning but lack necessary arguments.

The majority of these errors stem from predicate incompleteness, and specifically missing objects and complements. Additionally, we can see that the complex syntactic constructions, including coordination, appositions, and relative/adverbial clauses, account for the majority of structural failures in sentence decomposition. The gold standard set also contains more verbs and tokens per atomic sentence, which suggests that the rules-based system can produce results that are underspecified, or lacking important information from the original sentence.

These findings demonstrate that a rules-based system can produce atomic sentences reasonably well, but the vast variation in human language, and the ambiguity of certain structures means it is difficult for dependency-based methods to differentiate separate ideas when they seem to be presented syntactically as part of the same unit.

\section{Conclusion and Future Work}

\tab This study focused on examining how different syntactic structures affect the performance of a rule-based atomic sentence extraction system. By evaluating the system on 100 manually annotated sentences from the WikiSplit dataset, we found that the rule-based methods can reliably handle simple subject–verb–object constructions, but tend to struggle more with complex forms, such as relative clauses, appositions, coordinated predicates, and adverbial clauses. These structures often led to missing subjects or objects or just incorrect clauses, highlighting how syntactic complexity directly contributes to predicate incompleteness and boundary errors. Despite these limitations, the system achieved moderate performance metrics with the gold standard, demonstrating that rule-based extraction allows for a more transparent and interpretable solution while still preserving much of the original meaning.

The rule-based atomic sentence extraction system was evaluated against the previously discussed manually annotated gold-standard set of 100 WikiSplit source sentences, which produced 252 aligned atomic sentence pairs. Performance was measured using ROUGE and BERTScore, all of which together work to evaluate the lexical, structural, and semantic similarity of the sentence pairs. These results suggest that while lexical and structural overlap are relatively strong, there is still room to improve semantic completeness, particularly in cases involving embedded or coordinated clauses.

Rule‑based clause splitting has already been shown to benefit information extraction tasks, for example in pharmacological text mining where a linguistic clause‑splitting algorithm improved the extraction of drug–drug interactions from complex biomedical sentences (Segura‑Bedmar et al., 2011). Building on this, future work could explore how atomic sentence extraction performs in more specific applications such as question answering and fact verification. In addition, expanding and refining the rule set to better handle nested clauses, coordination, and appositions, or combining rules with components to recover missing arguments, could improve the accuracy of the overall system.

\section*{References}

\tab Botha, J. A., Faruqui, M., Alex, B., \& Ganchev, K. (2018). Learning to split and rephrase from Wikipedia edit history. arXiv. https://arxiv.org/abs/1808.09468\\

Liu, J., Cohen, S. B., \& Lapata, M. (2018). Discourse representation structure parsing. Transactions of the Association for Computational Linguistics, 6, 147–160. https://doi.org/10.1162/tacl\_a\_00011\\

Narayan, S., Gardent, C., Shimorina, A., \& Perez-Beltrachini, L. (2017). Split and rephrase. Proceedings of the 2017 Conference on Empirical Methods in Natural Language Processing (EMNLP), 606–616. https://doi.org/10.18653/v1/D17-1063\\

Niklaus, C., Cetto, M., Freitas, A., \& Handschuh, S. (2018). A survey on open information extraction. Proceedings of the 27th International Conference on Computational Linguistics, 3866–3878. https://aclanthology.org/C18-1326\\

Zhu, Z., Bernhard, D., \& Gurevych, I. (2010). A monolingual tree-based translation model for sentence simplification. Proceedings of the 23rd International Conference on Computational Linguistics, 1353–1361. https://aclanthology.org/C10-1158\\

Siddharthan, A. (2006). Syntactic simplification and text cohesion. Research on Language and Computation, 4, 77–109. https://doi.org/10.1007/s11168-006-9011-1\\

Zhang, X., \& Lapata, M. (2017). Sentence simplification with deep reinforcement learning. Proceedings of the 2017 Conference on Empirical Methods in Natural Language Processing, 584–594. https://doi.org/10.18653/v1/D17-1062\\

Chatterjee, N., \& Agarwal, R. (2021). DEPSYM: A lightweight syntactic text simplification approach using dependency trees. Proceedings of the Workshop on Computational Text Simplification (CTTS).\\

Kim, J., Maddela, M., Kriz, R., Xu, W., \& Callison-Burch, C. (2021). BiSECT: Learning to split and rephrase sentences with bitexts. Proceedings of the 2021 Conference on Empirical Methods in Natural Language Processing, 6193–6209. https://doi.org/10.18653/v1/2021.emnlp-main.500\\

Segura-Bedmar, I., Martínez, P., \& Sánchez-Cisneros, D. (2011). A linguistic rule-based approach to extract drug-drug interactions from pharmacological documents. BMC Bioinformatics, 12(Suppl 2), S1. https://doi.org/10.1186/1471-2105-12-S2-S1

\end{document}